\documentclass[twocolumn]{article}
\usepackage[utf8]{inputenc}

\usepackage{graphicx}

\usepackage{nccmath}
\usepackage[numbers]{natbib}

\usepackage{amsmath,amssymb,amsfonts}
\usepackage{algorithmic}
\usepackage{graphicx}
\usepackage{textcomp}
\usepackage{xcolor}
\usepackage{colortbl}
\usepackage{balance}
\usepackage{multirow}
\usepackage[utf8]{inputenc}
\usepackage[hyphens]{url}
\usepackage{hyperref}
\usepackage{subfigure}
\usepackage{comment}
\usepackage{balance}

\usepackage{fancyhdr}
\graphicspath{{./assets/}} 

\title{A Comparison of Humanoid Robot Simulators:\\A Quantitative Approach}
\author{
Angel Ayala\,$^{1}$ \and 
Francisco Cruz\,$^{2,3}$ \and
Diego Campos\,$^{2}$ \and
Rodrigo Rubio\,$^{2}$ \and
Bruno Fernandes\,$^{1}$ \and
Richard Dazeley\,$^{3}$
}

\date{\normalsize
    $^{1}$Escola Polit\'ecnica de Pernambuco, Universidade de Pernambuco, Recife, Brasil\\
    $^{2}$Escuela de Ingenier\'ia, Universidad Central de Chile, Santiago, Chile.\\
    $^{3}$School of Information Technology, Deakin University, Geelong, Australia.\\
    Corresponding e-mails: \{aaam, bjtf\}@ecomp.poli.br, \{francisco.cruz, richard.dazeley\}@deakin.edu.au, \\ \{diego.campos, rodrigo.rubio\}@alumnos.ucentral.cl\\
}

\begin{document}

\maketitle

\begin{abstract}
Research on humanoid robotic systems involves a considerable amount of computational resources, not only for the involved design but also for its development and subsequent implementation.
For robotic systems to be implemented in real-world scenarios, in several situations, it is preferred to develop and test them under controlled environments in order to reduce the risk of errors and unexpected behavior.
In this regard, a more accessible and efficient alternative is to implement the environment using robotic simulation tools.
This paper presents a quantitative comparison of Gazebo, Webots, and V-REP, three simulators widely used by the research community to develop robotic systems.
To compare the performance of these three simulators, elements such as CPU, memory footprint, and disk access are used to measure and compare them to each other.
In order to measure the use of resources, each simulator executes 20 times a robotic scenario composed by a NAO robot that must navigate to a goal position avoiding a specific obstacle.
In general terms, our results show that Webots is the simulator with the lowest use of resources, followed by V-REP, which has advantages over Gazebo, mainly because of the CPU use.
\end{abstract}


\thispagestyle{fancy}

\section{Introduction}
The implementation of robotic solutions represents a costly and time-consuming process.
For this reason, robot simulators have emerged as an important complementary tool~\cite{Andrews2012:Simulation}, being a fundamental part of the development of robotic solutions. 
Robot simulators allow evaluating the feasibility and efficiency of algorithms varying in type and complexity, in a more controlled environment with no disturbances avoiding the occurrence of accidents~\cite{cruz2016learning}.

In recent years, the number of simulation tools available has grown substantially for the use of different kinds of robots~\cite{Erez2015:EngineSimulation}.
For the use of humanoid robots, the development of novel solutions represents a complex challenge in the simulation due to the high number of joints, and the contact between different surfaces and textures~\cite{cruz2017agent, barros2020moody}.
In this regard, to evaluate available robot simulators, it is an important aspect for the scientific and academic community in order to facilitate the selection of the most suitable simulation tool~\cite{Ivaldi2014:SimToolsSurvey}.

\begin{figure}
    \centering
        \includegraphics[width=\columnwidth]{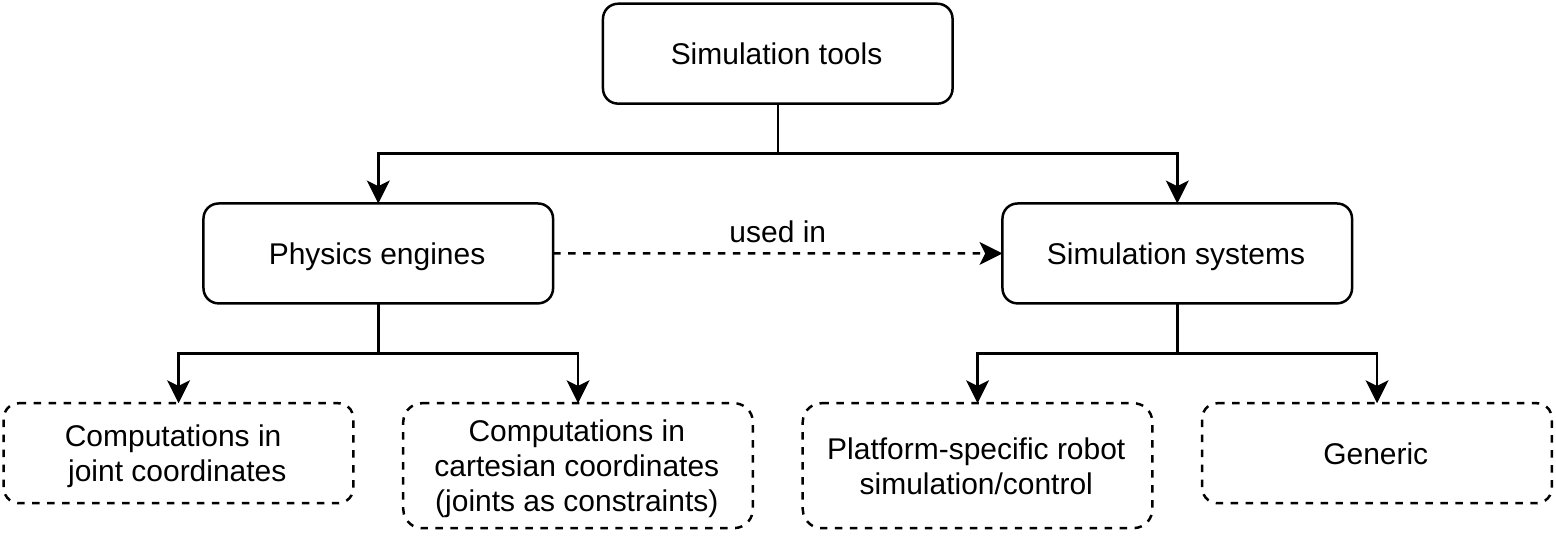}
        \caption{A simple classification of simulation tools, separated in two main categories; (a) Physics engines: are responsible for the representation of rigid-body structures and its dynamics, computed in joint coordinates (e.g., MuJoCo, XDE) or in cartesian coordinates (e.g., ODE); and (b) Simulation systems: as an integral environment for the development of simulations with a user interface, build upon a physics engine, platform-specific (e.g., iCub, HRP) or generic (e.g., V-REP, Gazebo).
        Adapted from~\cite{Ivaldi2014:SimToolsSurvey}.}
    \label{fig:SimToolsClass}
\end{figure}

In developmental robotics, robot simulators have been also widely used in order to simplify experimental analysis~\cite{tikhanoff2010integration, cruz2018action, moreira2020deep}. 
However, the decision about what simulator to use rarely consider aspects such as performance or use of resources. 
Previous works have addressed the comparison of simulators from different perspectives or points of view, including the comparison between different physics engines~\cite{Erez2015:EngineSimulation} and simulation systems~\cite{Ivaldi2014:SimToolsSurvey}.
Figure \ref{fig:SimToolsClass} presents a general overview of the previously evaluated simulation tools.
In this classification, a physics engine is considered the software responsible for representing rigid-body structures, as well as the dynamics of movement and contact between different structures.
A simulation system is a complete suite containing physics engines, which can incorporate a model editor, sensor simulation, and the possibility of interaction with the user.

A simulation tool must be able to compute a large number of mathematical operations to carry out the best possible reproduction of the real-world aspects in a computational way.
To perform a simulation task requires a determined amount of resources, and thus, different simulators can offer the execution of the implemented task with different configurations of resources depending on their working method.
Tasks that require a large number of calculations are those that must be fulfilled by humanoid robots, in part due to the number of joints that must be operated during a simulation~\cite{craighead2007survey}.

For robot simulators, there has been previous analysis and comparisons considering qualitative elements, covering hardly quantitative aspects.
The discussion of these comparisons is in the next section.
In this paper, we present a quantitative comparison approach for the Gazebo\footnote{See http://gazebosim.org/}, Webots\footnote{\label{label:webots}Since December 2018, Webots is released as an open source software under the Apache 2.0 license. See https://www.cyberbotics.com/}, and V-REP\footnote{\label{note:vrep}Since November 2019, V-REP simulator has been replaced by CoppeliaSim version 4.0. See https://www.coppeliarobotics.com/} simulators.
This comparison considers aspects such as CPU, memory footprint, and disk usage being evaluated using a humanoid robot scenario in a domestic situation.

\section{Related Works}
It is not simple to define a single aspect or metric to establish the best simulator.
Nevertheless, different authors have previously addressed qualitative aspects, and some of them have included quantitative aspects, focusing on comparing between open-source and closed-source simulation tools.
Ivaldi et al.~\cite{Ivaldi2014:SimToolsSurvey}, addressed a qualitative comparison of different simulation tools through a survey applied to 119 users.
Most of the users worked in control and locomotion with humanoid and mobile robots, doing research in academic, public, or private areas.
Additionally, experiments were carried out with the iCub humanoid robot in Gazebo, XDE, and the official iCub simulator, showing that the first two were capable of simulating contact more similarly to real-world.
The obtained results demonstrated that users prioritize a more realistic simulation and the use of the same code for simulated and real robots.
The Gazebo simulator was considered as the best choice of open-source software, and V-REP as the best commercial simulator with a free educational license.

A similar approach was addressed by Torres-Torriti et al.~\cite{Torres2016:SimSurvey}, who performed a comparative study of free simulation tools for mobile robots.
In their work, the authors presented a qualitative and quantitative comparison of three free publicly available simulation software: Carmen, Player-Stage-Gazebo, and Microsoft Robotics Developer Studio simulators, along with the Open Dynamics Engine (ODE) physics engine.
The selection of these tools was based on existing documentation and support as well as the time required for physical accuracy. 
In the qualitative comparison, the main physics modeling capabilities and middleware functionality were considered but without the user interface, due to it not being relevant to the use of the simulator.
For the quantitative comparison, two scenarios were considered to evaluate the simulations against the real-world experimental metrics. 
The first scenario considered longitudinal motion and the second one an open-loop control command sequence.
As a general conclusion, they did not claim any software as being superior to others. 
However, they showed that the ODE engine presented a more consistent simulation.

A recent work, presented by Pitonakova et al.~\cite{Pitonakova2018:SimComparison}, compared the V-REP, Gazebo, and ARGoS simulators.
For the qualitative aspects, the built-in features, robot libraries, programming methods, and the interface's usability were considered.
For the quantitative comparison, two benchmarks were defined. 
The first benchmark considered the execution through its graphical user interface, and the second one considered a headless execution, i.e., without the user interface.
Additionally, two scenarios had been set up; the first scenario considered a large 2D plane, and the second one an industrial building model with 41,600 vertices.
Each benchmark was executed with 1, 5, 10, and 50 robots in each scenario, defining three performance metrics: real-time factor ($sim_t / real_t$), CPU, and memory usage.
The results showed that the V-REP simulator was the most resource consuming, but at the same time presented the most significant amount of features, as well as the ability to create new threads to make efficient use of the CPU.
Moreover, ARGoS presented a better balance between robot quantity and physics accuracy, being a suitable option for simulation of swarm robotics tasks. 
However, in ARGoS, all models used had to be previously developed in OpenGL\footnote{OpenGL is an Application Programming Interface (API) to produce and build 2D and 3D objects.}.
In the study, Gazebo was shown to be slightly more efficient than ARGoS and with similar features as V-REP.
However, it was evaluated as being not very user-friendly.
Furthermore, other recent works also addressed comparisons of simulators under other concepts such as multi-robots systems~\cite{Noori2017:MultiRobotsSim}, agent-based simulators~\cite{Abar2017:AgentBasedSim}, and swarm robotics~\cite{Zhong2018:SwarmComparison}, among others.



\section{Simulation Software}
Simulation tools have allowed the rapid development of prototypes in controlled environments against possible failures, contributing to their implementation in the physical world~\cite{cruz2018multi}.
Furthermore, they make it possible to learn and study the physics of dynamic systems, as well as promoting collaboration with the scientific community through the exchange of early knowledge.

\begin{figure*}
  \centering
  \subfigure[Gazebo simulator.
  ]{\includegraphics[width=0.31\textwidth]{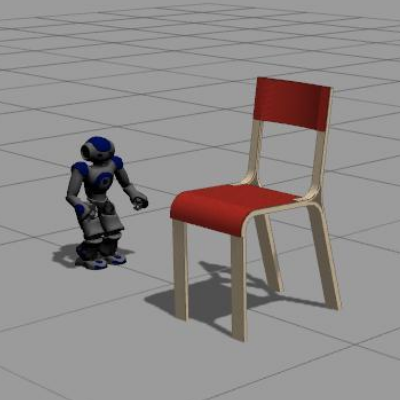} \label{fig:NAOGazebo}
  }\quad
  \subfigure[Webots simulator.
  ]{\includegraphics[width=0.31\textwidth]{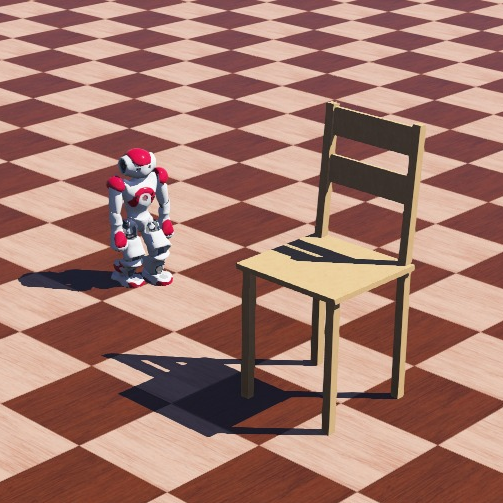} \label{fig:NAOWebots}
  }\quad
  \subfigure[V-REP simulator.
  ]{\includegraphics[width=0.31\textwidth]{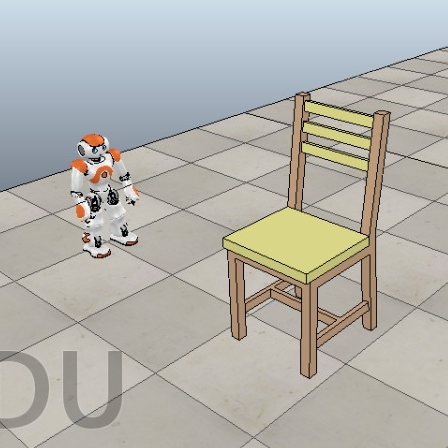} \label{fig:NAOCoppeliaSim}}
  
  \caption{The robotic scenario for the evaluation of the different simulators.
  This scenario is composed of a NAO robot and a chair.
  The NAO has to navigate around the chair, avoiding a collision, to reach the goal on the other side.} 
  \label{fig:Environment}
\end{figure*}

In this regard, simulation has become an essential part in developing robotic solutions, contributing to optimize algorithms for motion, rapid prototyping of controllers and actuators, as well as prior environment verification.
Nowadays, there are a large number of software tools for simulations that are available for the scientific community. 
However, up to now, comparisons in terms of computational resources considering humanoid robots have been only briefly covered.
The main difficulty in analysing and comparing simulators is that they have different requirements and characteristics.
Therefore, the robot performance cannot always be evaluated directly in a simulation environment~\cite{Ivaldi2014:SimToolsSurvey, Koenig2004:Gazebo}.

In order to get a computer simulation, a physics engine is required~\cite{Erez2015:EngineSimulation}.
The physic engine is the software capable of modeling different dynamic systems of physics such as particles or contact, among others.
Afterward, it is common to use a development environment software to facilitate the simulation development, where the elements of the simulation are defined~\cite{Ivaldi2014:SimToolsSurvey}.
Another important aspect that has been developed over the last few years is the use of the Robot Operating System (ROS), a flexible framework created in 2007 by Stanford's artificial intelligence laboratory for the development of software for robots~\cite{quigley2009ros}.
ROS is a middleware that provides a large number of tools and libraries that aim to simplify the task of creating a complex and robust robotic behavior in a wide variety of robotic platforms.
A great advantage of ROS is the possibility of using the developed code for both the simulated and the real environment.

The simulation software selected for this work are Gazebo, Webots, and V-REP, having received the most positive evaluations and are the most widely used open-source (Gazebo and Webots) and closed-source (V-REP) as mentioned in~\cite{Ivaldi2014:SimToolsSurvey}.
Additionally, the physics engine used to evaluate the different simulation systems corresponds to the Open Dynamics Engine (ODE\footnote{See https://www.ode.org/}) that has been previously analyzed and shown the most physical consistency~\cite{Torres2016:SimSurvey, Pitonakova2018:SimComparison}.
All the simulation systems in this study are compatible with ROS for the execution of the simulations, moreover, most of them can be used through C/C++ language, JAVA, Python, among others. 
Below is a description of each simulator.

\subsection{Gazebo}
Gazebo's development has been driven by the increasing use of robotic vehicles~\cite{Noori2017:MultiRobotsSim}. 
It is capable of simulating interactions between robots in indoor and outdoor environments, providing realistic sensor feedback. 
Gazebo is thus designed to accurately reproduce the dynamics of the environments that a robot may encounter. 
This simulator work with two processes, client and server, which is capable of simulating from a remote machine.
Furthermore, it is entirely open-source and freely accessible, with a broad base of contributors supporting this tool. 
In addition, the default physics engines integrated with Gazebo include ODE, Bullet, Simbody, and DART~\cite{Koenig2004:Gazebo}.

\subsection{Webots}
Webots is an open-source robot simulator that provides a complete development environment~\cite{michel2004cyberbotics} for modeling, programming, and simulating robots~\cite{Shahverdi2016:NAO}.
Thanks to its easy to use and friendly interface, it can add or remove objects or robots and evaluate their possible benefit of the simulation scenario, requiring a small amount of time for development.
Moreover, it includes a compiler, which makes possible to test and validate control algorithms involving complex data processing quickly. 
Like Gazebo and V-REP, it comes with the ODE physical engine integrated.

\subsection{V-REP}
V-REP was introduced as a versatile and scalable simulation framework. 
By offering a multitude of different programming languages, it allows embedding controllers and functionality into simulation models, which facilitates the task for developers and reduces implementation complexity for the users~\cite{Rohmer2013:VREP}. 
It has now grown into a robust and widely used robot simulator, available for both academically and in the industrial field.
V-REP is closed-source with a free educational license. 
Moreover, V-REP has different options for physics engines, including Bullet Physics, ODE, Newton, and Vortex Dynamics. 


\section{Robotic Scenario}
The defined scenario used for the quantitative evaluation of the simulators involves a task performed by a humanoid robot.
The utilized humanoid is a NAO robot~\cite{Gouaillier2009:NAOMechatronic}, which has to perform the task of navigating to a goal position, avoiding an obstacle located between its starting position and the goal. 
This object on the scene is a chair that remains motionless at all times.
In this scenario, the humanoid robot must be able to go around the object in order to avoid the collision and complete the task.
In Figure \ref{fig:Environment}, the scenario is shown in the different simulators.
The definition of such a scenario is mainly based on evaluating the performance under a large number of turns and movements made by the robot, thus requiring the use of multiple joints.
The execution of multiple movements by the robot makes it possible to observe the workload required for each simulator to be capable of performing the task.
To estimate the workload, CPU, memory, and disk use metrics are used for each simulation run.
No other software has been executed during the simulation to prevent the measurement be affected by other factors.
\subsection{NAO Robot}
NAO is a humanoid robot developed in 2008 by Aldebaran Robotics, a French company subsidiary of the Softbank Robotics Group. 
It is primarily used for education and research of humanoid robots.
Among its main features is the ability to perceive the environment from its multiple sensors, including two cameras, four microphones, nine tactile sensors, two ultrasonic sensors, eight pressure sensors, an accelerometer, and a gyroscope. 
Moreover, it includes other expression elements that give it a high degree of interactivity, like its 53 RGB LEDs, its voice synthesizer, and its two speakers~\cite{Shahverdi2016:NAO}.
The software structure is based on the open-source Linux operating system and supports programming languages such as C, C++, URBI, Python, and .NET Framework. 
Additionally, a graphical interface has been developed for the robot called Choreographe~\cite{Pot2009:Choregraphe}.
Choreographe allows interactive programming of actions with different levels of complexity using flow diagrams.
This interface provides the ability to work in line with the robot hardware, maintaining dialogues and even obtaining object and person recognition through its cameras. 
This software is compatible with different operating systems such as Windows, iOS, and GNU/Linux.
Moreover, NAO has native support in Webots and unofficial support in V-REP and Gazebo simulators.

\subsection{Metrics software}
To measure the resources utilized during the simulation two tools have been used:

\begin{itemize}
    \item \textbf{GNU Monitor System} to measure in real-time the performance of our machine (see section IV.C). 
    This application is used to obtain information of the performance associated to three different aspects: (i) processes, presenting a list with all the processes in execution and capable of ordering them according to the resource they are using at every moment; (ii) resources, it is possible to visualize a real-time graph of memory and CPU usage; and (iii) file system, it is possible to obtain information regarding the space occupied on the hard disk.
    
    \item \textbf{Iotop Monitor} is a free, open-source utility similar to the top command, which provides an easy way to monitoring disk usage details through a table of existing usage per process or sub-process in the system.
    The Iotop tool is developed in Python and requires the kernel counting function to monitor and display the processes.
    It is a handy tool for system administrators to track specific processes that may cause a high level of reading/writing on the disk.
\end{itemize}

\subsection{Machine specifications}
The comparison was carried out using a virtual machine configured in VirtualBox v5.2 comprising an Intel\textsuperscript{\textregistered} Core\textsuperscript{\texttrademark} i5-2410M @2.30GHz, 6GB of memory, and 320GB  Hitachi SATA II @7200 RPM hard disk drive. 
The selected operative system for this purpose was Ubuntu 18.04.1 LTS, installed only with the essential packages required for its execution.
The Ubuntu distribution and the version chosen presents a wide support community for the installation and configuration of different simulation tools, as well as great compatibility with ROS.

\section{Simulation and Results}
The task introduced in the previous section was implemented to achieve the measurement of the quantitative data.
To code the robot's behavior, Choreographe was used for V-REP and Gazebo simulators.
For Webots, internal library methods were directly employed to produce the same robot's behavior.
This introduces no significant difference in the experiments since the metrics were measured isolated for each simulator, using Choreographe just as a graphical tool to design the robot's movement.
As previously discussed, we have defined as comparative elements for the different simulators, the use of CPU, memory footprint, and disk access.
To specify the impact of each defined comparative element over the global comparison, these were weighted equally, showing no predisposition to favor any particular aspect.

\begin{figure}[t]
    \centering
    \includegraphics[width=\columnwidth]{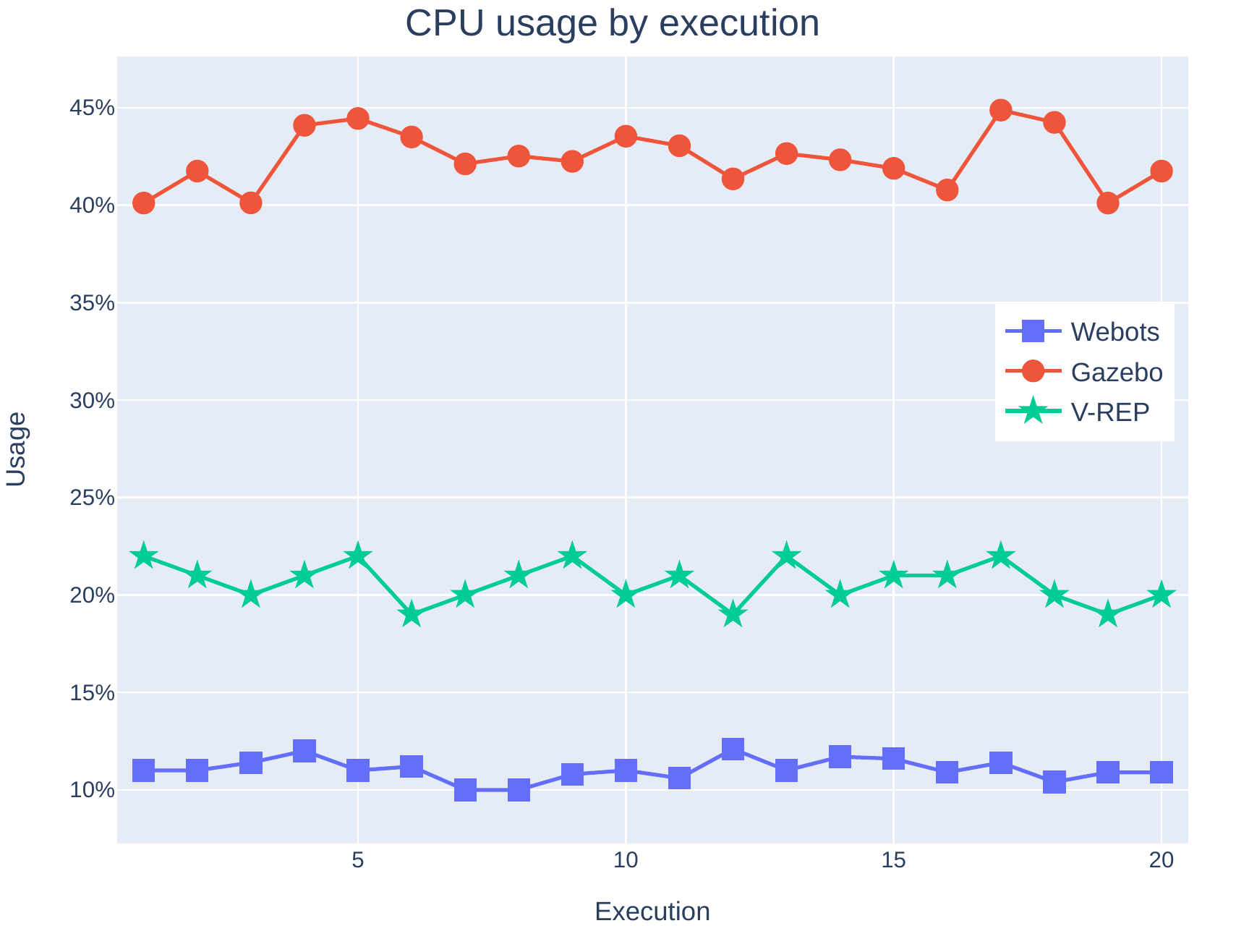}
    \caption{The CPU required for each simulator during 20 executions of the robotic scenario.
    The Webots simulator required about an 11.05\% of CPU to execute the scenario, followed by V-REP with an average of 20.65\%.
    Finally, Gazebo required considerable more CPU (42.38\% average) to simulate the task.}
    \label{fig:CPUPerformance}
\end{figure}

To compare the performance, the simulated scenario was run 20 times for each simulator.
In Figure \ref{fig:CPUPerformance} is shown the CPU load for each simulation run.
The Webots simulator required the lowest amount of processing power, using in average 11.05\% of the processor to execute the simulation tasks.
Following, V-REP used almost twice of CPU, with an average of 20.65\%.
For Gazebo, the use of CPU was the highest, presenting an average of 42.38\%, for the execution of a simple simulation task.
The issue that Gazebo used this amount of CPU is attributed to the use of two processes: gzclient (client) and gzserver (server).

\begin{figure}[t]
    \centering
    \includegraphics[width=\columnwidth]{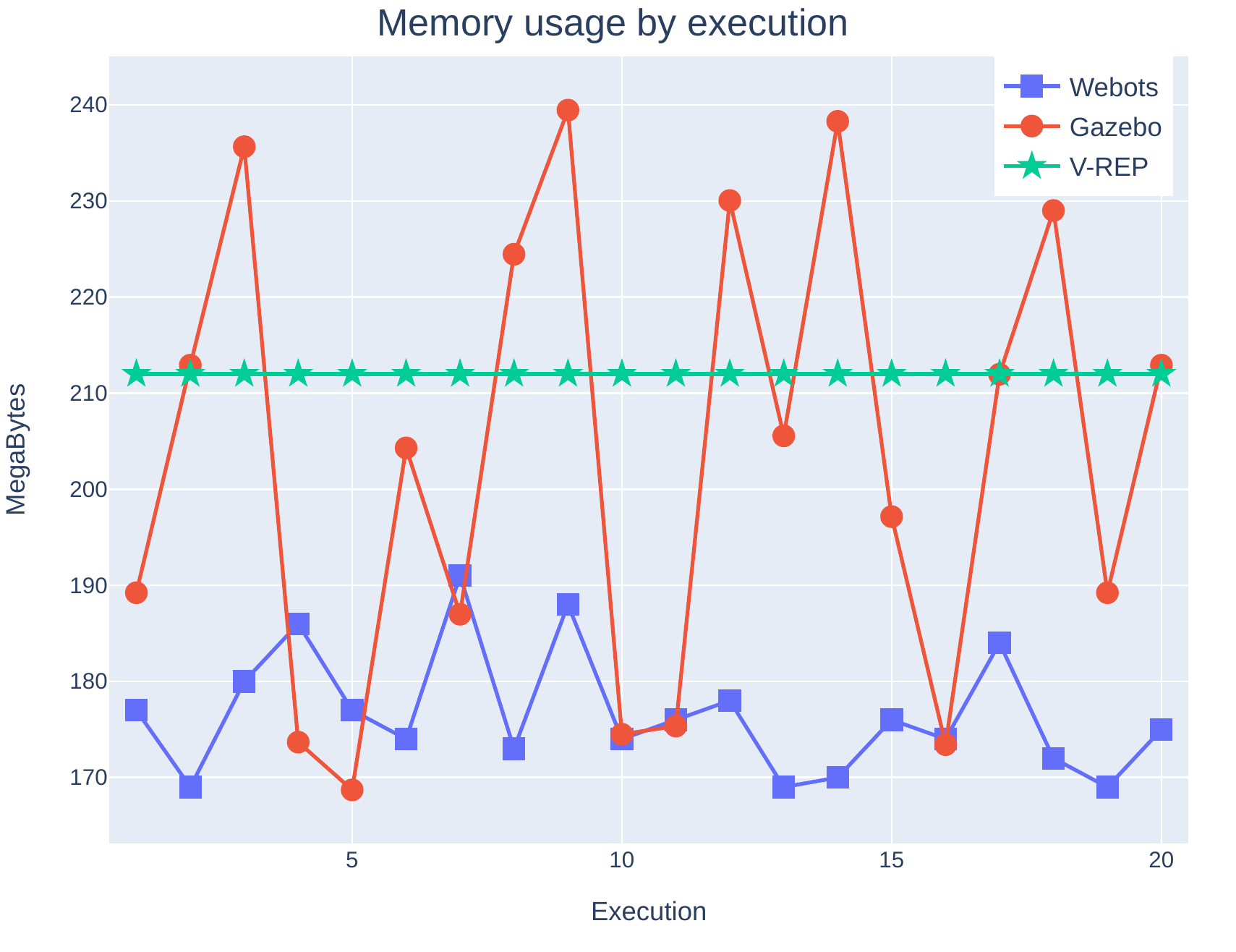}
    \caption{The memory required for each simulator during 20 executions of the robotic scenario.
    The V-REP simulator presented a stable memory use (just 212MB), moreover, the Webots simulator required less memory (176.6MB average), followed by Gazebo with 203.64MB average for the task execution.}
    \label{fig:MemoryPerformance}
\end{figure}

The memory footprint, in Figure \ref{fig:MemoryPerformance}, is represented in Megabytes required for the simulation execution.
Firstly, the memory footprint of V-REP was stable, always requiring the same amount of memory.
Quite the opposite was observed in the case of Webots and Gazebo, requiring a variable amount of memory for each execution.
Webots presented a more stable memory footprint with low differences between each execution, while Gazebo showed a significant memory difference between executions.

\begin{figure}[t]
    \centering
    \includegraphics[width=\columnwidth]{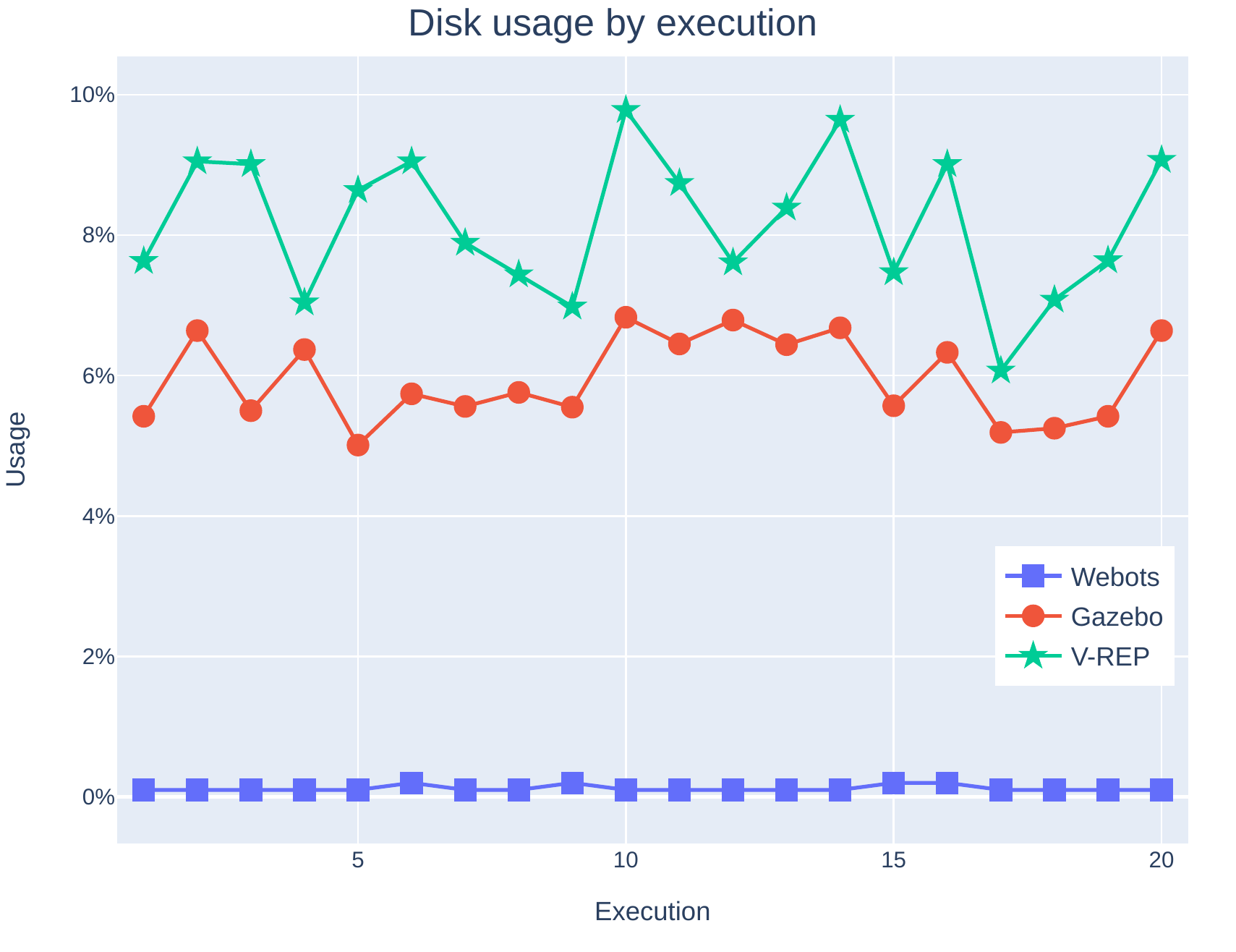}
    \caption{The disk use for each simulator during 20 executions of the robotic scenario.
    It can be observed that Webots required almost no disk access (0.12\% average) for the execution of the simulation.
    For Gazebo, an average of 5.96\% disk access is required, followed by V-REP with 8.16\%.}
    \label{fig:DiskUse}
\end{figure}

The disk usage metric, in Figure \ref{fig:DiskUse}, presents the transactions that are allocated for disk access to write/read as percentage of all performed operations.
In general, all the simulators required a low amount of disk transactions.
For Webots, almost no disk use was required in order to execute the simulation, presenting a 0.12\% average of disk access.
Gazebo and V-REP required almost the same disk usage, with an average of 5.96\% and 8.16\%, respectively.

\begin{figure*}[t]
  \centering
  \subfigure[CPU use boxplot of 20 executions.
  The simulator with smallest standard deviation is Webots with 0.56, followed by V-REP with 1.04, and Gazebo with 1.46.
  ]{\includegraphics[width=0.31\textwidth]{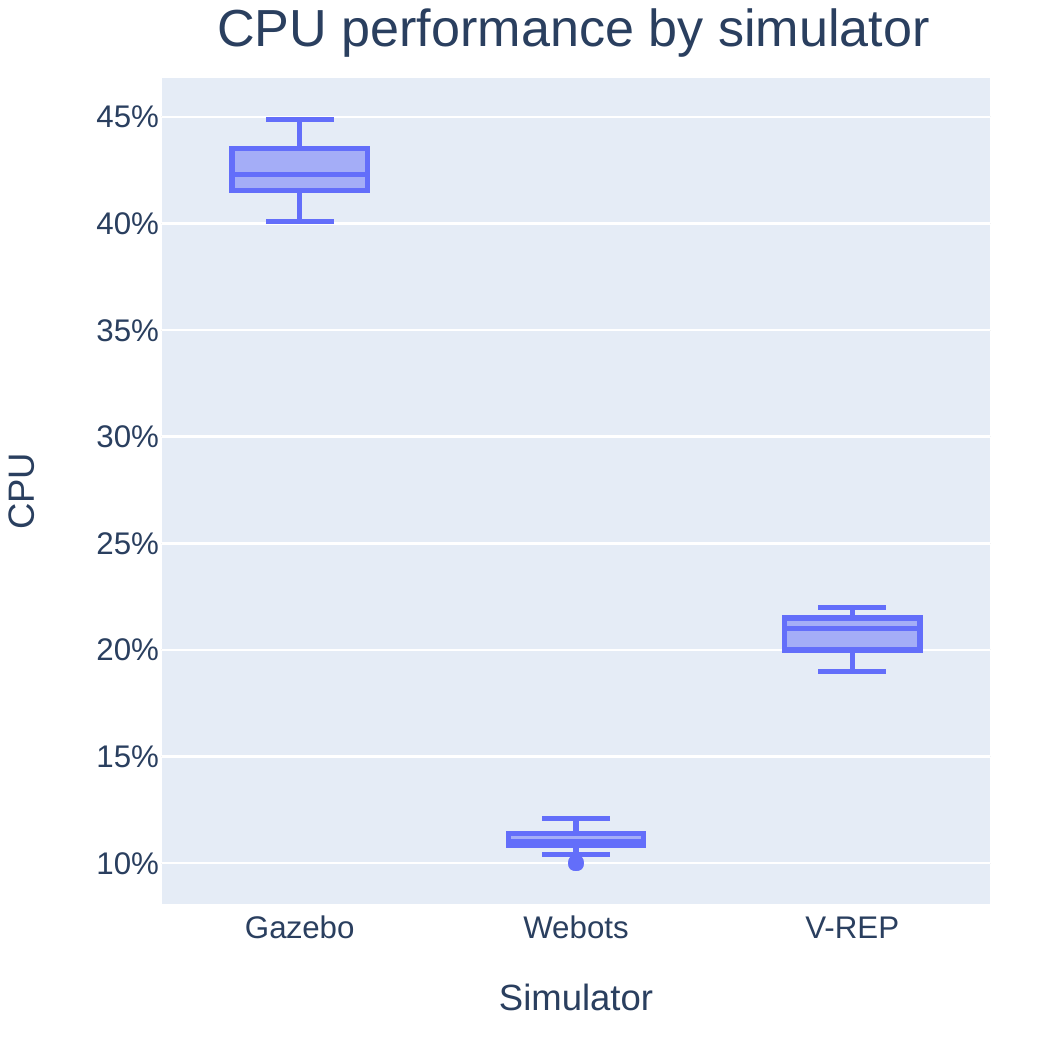} \label{fig:CPUBoxplot}
  }\quad
  \subfigure[Memory boxplot of 20 executions.
  The simulator with smallest standard deviation is V-REP with 0.0, followed by Webots with 6.36, and Gazebo with 24.0.
  ]{\includegraphics[width=0.31\textwidth]{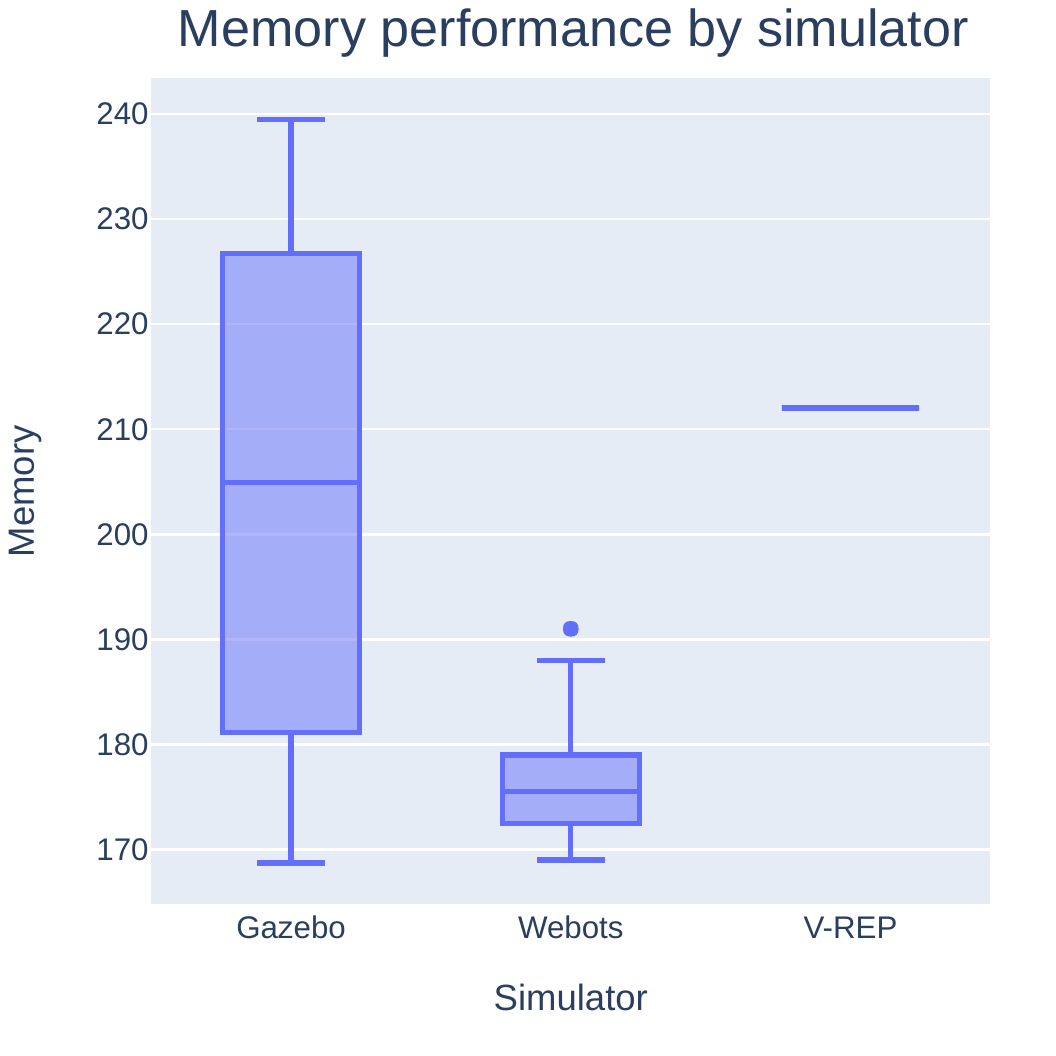} \label{fig:MemoryBoxplot}
  }\quad
  \subfigure[Disk access boxplot overall of 20 executions.
  The simulator with smallest standard deviation is Webots with 0.04, followed by Gazebo with 0.61, and V-REP with 1.01.
  ]{\includegraphics[width=0.31\textwidth]{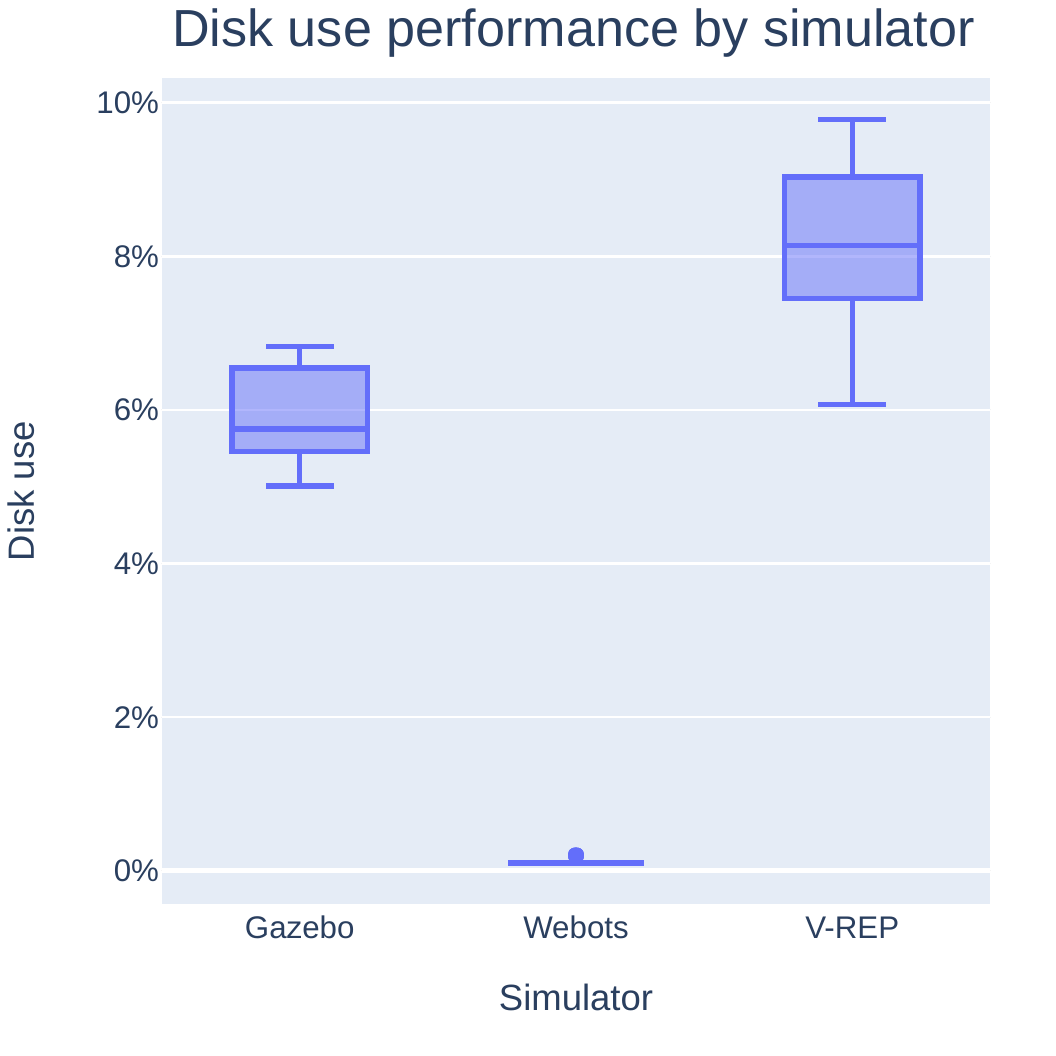} \label{fig:DiskBoxplot}
  }
  
  \caption{Averaged results for 20 executions of the measured elements to compare the simulators in a quantitative approach.
  Webots is the simulator that requires less amount of resources in order to execute the humanoid simulation task.
  Follow, the Gazebo simulator required fewer resources than V-REP, but this presented more CPU use, and the memory footprint presented significant differences between executions.
  Finally, the V-REP is the second simulator that used less amount of resources, with a stable memory consumption.}
  \label{fig:OverallBoxplot}
\end{figure*}

Moreover, Figure \ref{fig:OverallBoxplot} summarizes the results of the 20 executions through statistical boxplots representation for each comparative element and each simulator.
It can be seen that both CPU and disk usage, shows a considerable variation among simulators.
Generally speaking, Webots is the simulator that uses fewer resources for the task execution, being able to simulate efficiently with CPU use $\in[10\%; 12.1\%]$ and memory footprint $\in[169\text{MB}; 191\text{MB}]$.
Additionally, the simulation is executed in a fast way, given that no disk operations were required (values $\in[0.1\%; 0.2\%]$). 
In comparison, V-REP requires almost twice as much CPU than Webots, with CPU use $\in[19\%; 22\%]$, and the simulation execution performs some on-disk operations with values $\in[6.07\%; 9.78\%]$. 
However, the memory footprint presented for V-REP is stable over all the runs; this is an advantage, even being 20.39\% (on average) higher than the memory required by Webots.
Finally, Gazebo presents fewer on-disk operations than V-REP, with values $\in[5.01\%; 6.83\%]$. 
Nevertheless, due to the client and server processes required for the simulation execution, Gazebo uses twice as much CPU than V-REP, with CPU use $\in[40.11\%; 44.88\%]$.
In relation to the memory, although Gazebo presents a lower average amount than V-REP, also shows instability with great variations between executions.

\section{Conclusions}
Research in the area of robotics present a high computational cost in its development, from planning how the system works to hardware implementation.
To ensure that a certain solution is feasible to be implemented in a real-world scenario, in several occasions, it is executed and tested in a controlled and simulated environment.
In this work, simulation systems such as Gazebo, Webots, and V-REP have been compared, evaluating the use of resources such as CPU, memory footprint, and disk access.
For this evaluation, a task using the humanoid robot NAO has been implemented, in which the robot has to navigate to a goal position avoiding an obstacle in the middle of its trajectory.

The evaluation of the simulators, have shown similar characteristics among them, such as compatibility with a Linux-based operating system, as well as its modular communication capability through an API in programming languages such as Python and C/C++.
Additionally, they all include the ODE engine for the simulation of physics in the virtual environment.
Contrarily, all the simulators present different complexities for the development of the proposed simulated scenario.
In Webots and V-REP, NAO is natively integrated, while in Gazebo, it is necessary to make a manual installation with third-party libraries. 
Although the robot must be controlled by an external API, it has been shown through the obtained results to be quite effective, achieving a successful simulation without any significant drawbacks.

Analyzing the results, Webots obtains the best score to execute the simulation with the NAO robot, requiring the less amount of resources in order to perform the simulation task.
However, V-REP presents a stable use of memory with a large library of models for simulation.
Finally, Gazebo also requires almost the same resources, considering it contains two processes due to a networking implementation design, and without having the NAO model integrated.

The obtained results may inspire future research in different directions. 
Future work includes considering other metrics for the comparison, such as time, with different physics engines as well as the inclusion of other robot simulators.
Additionally, a more comprehensive comparison should take into account different kind of scenarios, from simple tasks to more complex ones using tools as direct planning, direct and inverse kinematics, collision avoidance, and image processing among others, as well as other kinds of robot platforms.
Moreover, a more fair comparison could weigh in a different way each quantitative aspect, giving more preponderance to some of them, as well as including qualitative aspects into the comparison for mixed approaches.

\section*{Acknowledgment}
This work has been financed in part by Universidad Central de Chile under the research project CIP2018009, the Coordenação de Aperfeiçoamento de Pessoal de Nível Superior - Brasil (CAPES) - Finance Code 001, Fundação de Amparo a Ciência e Tecnologia do Estado de Pernambuco (FACEPE), and Conselho Nacional de Desenvolvimento Científico e Tecnológico (CNPq) - Brazilian research agencies.


\bibliographystyle{ieeetr}
\balance
\bibliography{references}

\end{document}